\documentclass{article}
\usepackage{spconf,amsmath,epsfig}
\usepackage{siunitx}
\usepackage{pifont}
\usepackage{subfig}
\usepackage{booktabs} 

\usepackage{algpseudocode}
\usepackage{url} 
\usepackage{comment} 

\renewcommand{\paragraph}[1]{\noindent\textbf{#1}\quad}

\title{AUTOMATED NATIONAL URBAN MAP EXTRACTION}

\fourauthors
  {Hasan Nasrallah}
  {National Center for \\ Remote Sensing - CNRS \\Beirut, Lebanon}
  {Abed Ellatif Samhat}
  {Faculty of Engineering\\ Lebanese University\\ Hadath, Lebanon}
  {Cristiano Nattero}
  {WASDI\\ Luxembourg \\}
  {Ali J. Ghandour\sthanks{Correspondence: aghandour@cnrs.edu.lb.}}
  {National Center for \\ Remote Sensing - CNRS \\Beirut, Lebanon}
  
\begin{document}
%
\maketitle
\begin{abstract}
Developing countries usually lack the proper governance means to generate and regularly update a national rooftop map. Using traditional photogrammetry and surveying methods to produce a building map at the federal level is costly and time consuming. Using earth observation and deep learning methods, we can bridge this gap and propose an automated pipeline to fetch such national urban maps.
This paper aims to exploit the power of fully convolutional neural networks for multi-class buildings' instance segmentation to leverage high object-wise accuracy results. Buildings' instance segmentation from sub-meter high-resolution satellite images can be achieved with relatively high pixel-wise metric scores. We detail all engineering steps to replicate this work and ensure highly accurate results in dense and slum areas witnessed in regions that lack proper urban planning in the Global South. We applied a case study of the proposed pipeline to Lebanon and successfully produced the first comprehensive national building footprint map with approximately 1 Million units with an 84\% accuracy. The proposed architecture relies on advanced augmentation techniques to overcome dataset scarcity, which is often the case in developing countries.
\end{abstract}

\section{Introduction}

Building footprint extraction from aerial imagery is an important step for many urban applications.
Fully automated extraction and recognition of buildings footprint geometries can also be used for a wide range of scientific applications in various domains such as solar rooftop potential estimation, solid waste management, and air pollution modeling, among others. 

Classifying pixels into semantic and instance objects in urban areas satellite images is currently undergoing important attention in the research community, in addition to development efforts in the industry. Remote sensing images are complex data, characterized by the form of heterogeneous regions with large intra-class variations and often lower inter-class variations \cite{ALSHEHHI2017139}. Deep learning significantly reduces the time required to achieve such tasks because of its ability to automatically extract meaningful and well-defined features and patterns present in large scenes. 

Given the aforementioned time and cost advantage over other traditional methods, such as on-site measurements or pure image processing, many methods have been proposed to solve the problem of building segmentation from aerial imagery. In \cite{GGCNN}, the authors use gated graph convolutional neural networks to produce a transcribed signed distance map (TSDM), which is then converted into a semantic segmentation mask of buildings. In \cite{RA-FCN}, the authors propose two plug-and-play modules to generate spatial- and channel-augmented features for semantic segmentation from satellite images. In \cite{Sat+GIS} the authors use augmentations such as slicing, re-scaling, and rotations and additional GIS data to improve building footprint extraction. In \cite{siam_unet_damage}, the authors use siamese networks to segment and classify buildings present in pre- and post-disaster images. However, a more direct approach is presented in \cite{ternausv2} where the authors use only a semantic segmentation network with an additional output mask that designates the spacing between very close buildings to separate building instances. 

In this work, we introduce a multi-class high-accuracy segmentation architecture for extracting rooftop geometries from satellite images using deep convolutional neural networks. The contribution of this paper is threefold: \textit{ (i)} present a fully automated pipeline to extract urban map at the national level, \textit{(ii)} translate the problem at hand to a multi-class segmentation approach by adding two intermediate classes (border and spacing), and \textit{ (iii)} finally apply the proposed workflow to Lebanon as a case study.

\section{Methodology}
\label{method}

\subsection{Multi-class Segmentation}

\begin{figure*}[t]
\begin{center}
   \includegraphics[width=1.0\linewidth]{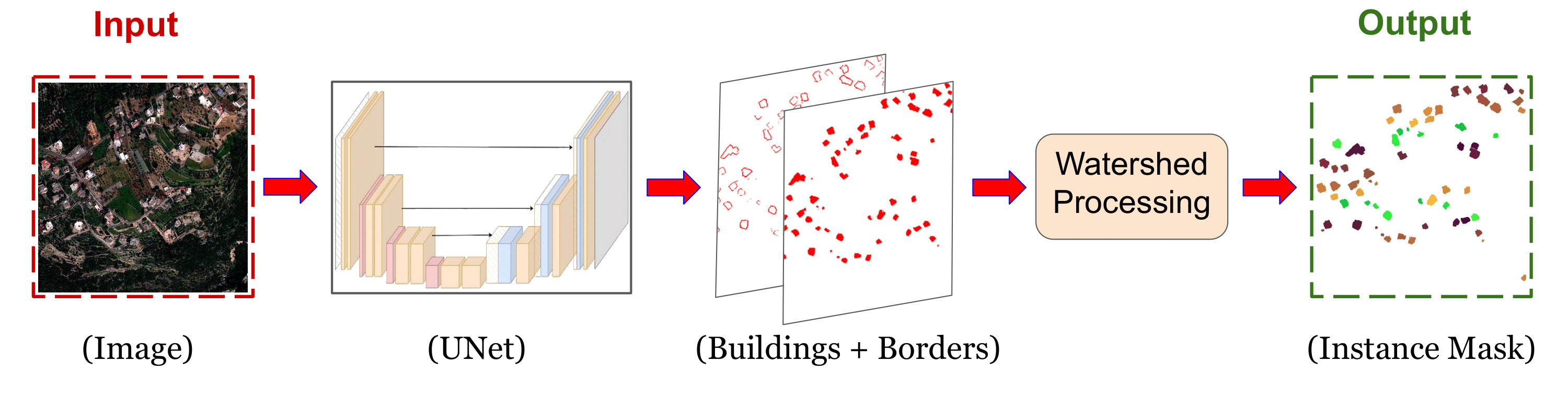}
\end{center}
   \caption{Proposed automated pipeline for national urban map extraction.}
\label{proposed_method}
\end{figure*}

In order to achieve high national wide segmentation score, especially in dense and slum areas, we define the following three output classes:

\begin{itemize}

\item \textbf{Building Class}: Pixels belonging to the interiors of any building polygon.

\item \textbf{Border Class}: Pixels belonging to the contours of the buildings. We create borders of width = 2 pixels.

\item \textbf{Spacing Class}: Pixels belonging to the separation between very close buildings.
\end{itemize}

Our best model is trained for 100 epochs using Adam Optimizer \cite{adam} and the One-Cycle learning rate policy \cite{oclr} using a Titan-Xp GPU. We used mixed precision training \cite{amp} and batch size = 16. During training, we apply random augmentations on images, including positional transformation, color, distortion, and noise transformations. Furthermore, to increase the robustness of the model, we use CutMix \cite{cutmix} data augmentation.

\subsection{Study Area}
\label{studyarea}

\begin{figure}[t]
\begin{center}
   \includegraphics[width=1.0\linewidth, height=1.05\linewidth]{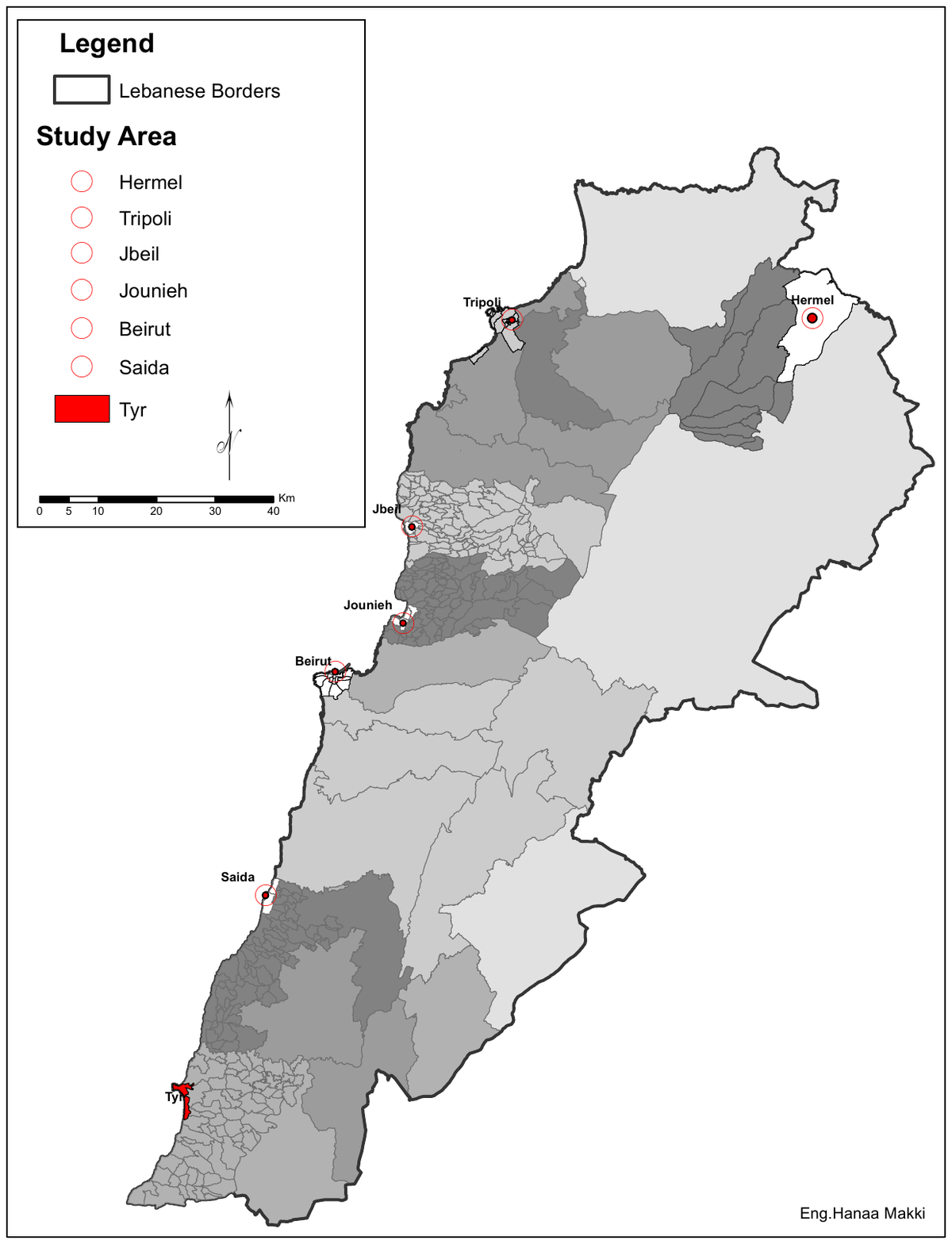}
\end{center}
   \caption{Study Area.}
\label{lebanon_map}
\end{figure}

\begin{figure}[t]
\begin{center}
   \includegraphics[width=1.0\linewidth]{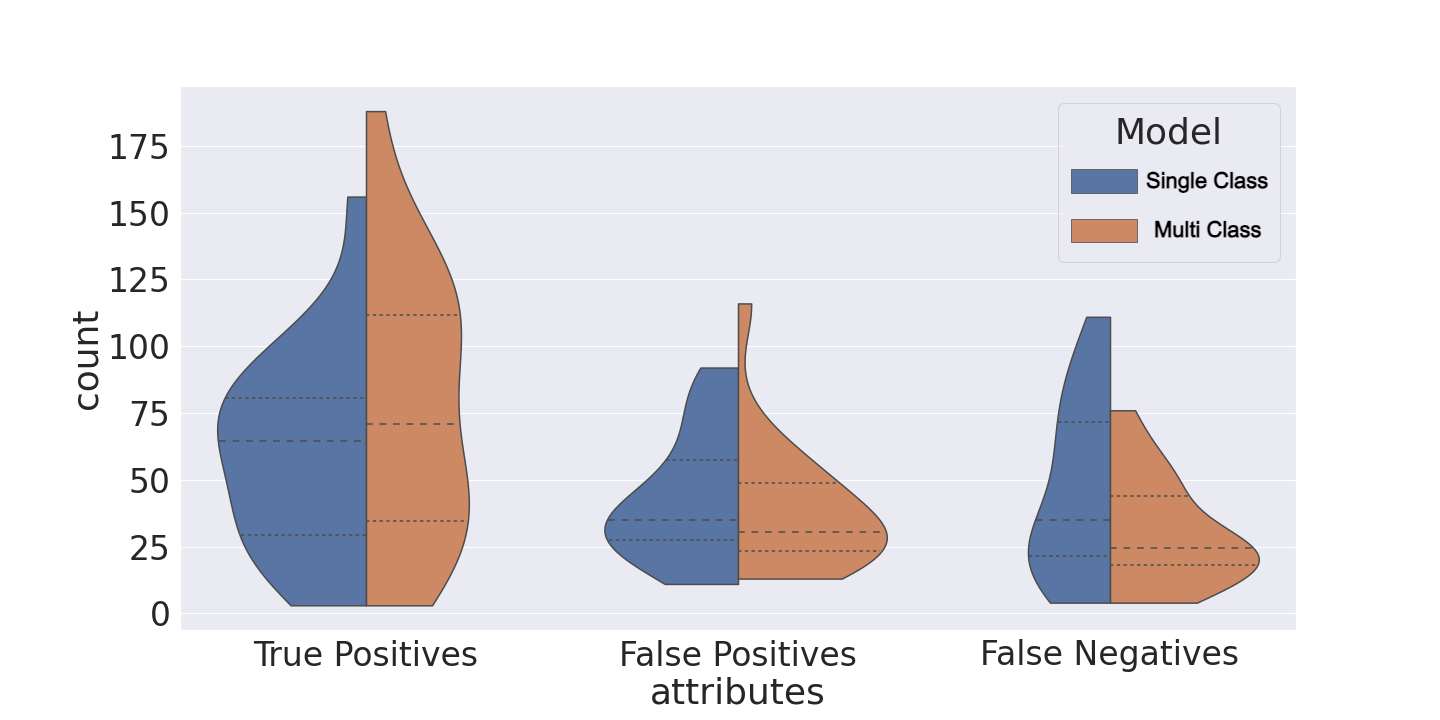}
\end{center}
   \caption{Violin Plots showing different metrics probability density distributions to compare the instance segmentation performance between the single and multi-class models.}
\label{violin}
\end{figure}

To create our dataset, we relied on a 49-cm spatial resolution RGB raster image that covers the city of Tyre and its neighborhood in the South of Lebanon with a total area of 255 kmsquare. The labeling process took about 100 hours and the final dataset is made up of 1,352 tiles of 512x512 and ~10,000 buildings. We split the data set into five folds and then automatically generate ground-truth border masks. To create the spacing mask between close buildings, we also followed several pre-processing techniques that will not be mentioned here for space limitation. Figure ~\ref{proposed_method} summarizes the pipeline proposed to extract urban map at the national level using multi-class instance segmentation by adding two intermediate classes (border and spacing).

As for the test set, we selected 30 Areas of Interest (AoI) of 1024x1024 size from different cities in Lebanon including Beirut, Saida, Jounieh, Jbeil and Tripoli. This would help to assess the generalizability of our model over the whole Lebanese geographical area. These AoI's contained dense and slum urban regions (Beirut and Tripoli), areas with proper urban planning (Saida), and some suburban and rural areas (Jbeil) as shown in Figure ~\ref{lebanon_map}.

\section{Experimental Results}
\label{resultss}

In our experiments, Efficient-Net-B3 was found to perform better in terms of accuracy and variance compared to other members of the Efficient-Net family members and other encoders such as ResNet34, ResNeXt50, InceptionV4, InceptionResNetV2 and DPN92. We also inspect the effect of mixed data augmentations such as Cutmix \cite{cutmix} and MixUp \cite{mixup}. When using Cutmix, the IoU score increased from 73.2\% to 75.1\% and the distance between the train and the validation scores decreases considerably. However, using Mixup does not improve model performance. Table \ref{tblearningrate} summarizes various combinations of hyperparameters experimented.

\begin{table*}[ht]
\begin{center}
\begin{tabular}{S|SSSS||S} \toprule
     {N} & {Backbone} & {AMP} & {Mixed Augs} & {Scheduler} & {Fscore} \\ \midrule
     1 & {EffNet-B2} & \ding{55} & \ding{55} & {PolyLr}  & {83} \\
     2 & {EffNet-B3} & \ding{51} & \ding{55} & {PolyLr}  & \textbf{84.3} \\
     3 & {EffNet-B4} & \ding{51} & \ding{55} & {PolyLr}  & {83.8} \\ \midrule
     4 & {EffNet-B3} & \ding{51} & {MixUp} & {PolyLr}  & {83.2} \\
     5 & {EffNet-B3} & \ding{51} & {CutMix} & {PolyLr}  & \textbf{85.6} \\ \midrule
     6 & {EffNet-B3} & \ding{51} & {CutMix} & {CycLR}  & {85.2} \\ 
     7 & {EffNet-B3} & \ding{51} & {CutMix} & {CycLR WM}  & {85.2} \\ 
     8 & {EffNet-B3} & \ding{51} & {CutMix} & {1Cycle}  & \textbf{87.6} \\ \bottomrule
     
\end{tabular}
\end{center}
\caption{Pixel Fscore results on our validation dataset showing a series of improvements done by searching for the best hyper-parameters like backbone architecture, automatic mixed precision, mixed augmentation and learning rate schedulers.} 
\label{tblearningrate}
\end{table*}

\begin{figure}[h]
\begin{center}
\subfloat[Saida Sample]{\includegraphics[width = 1.5in]{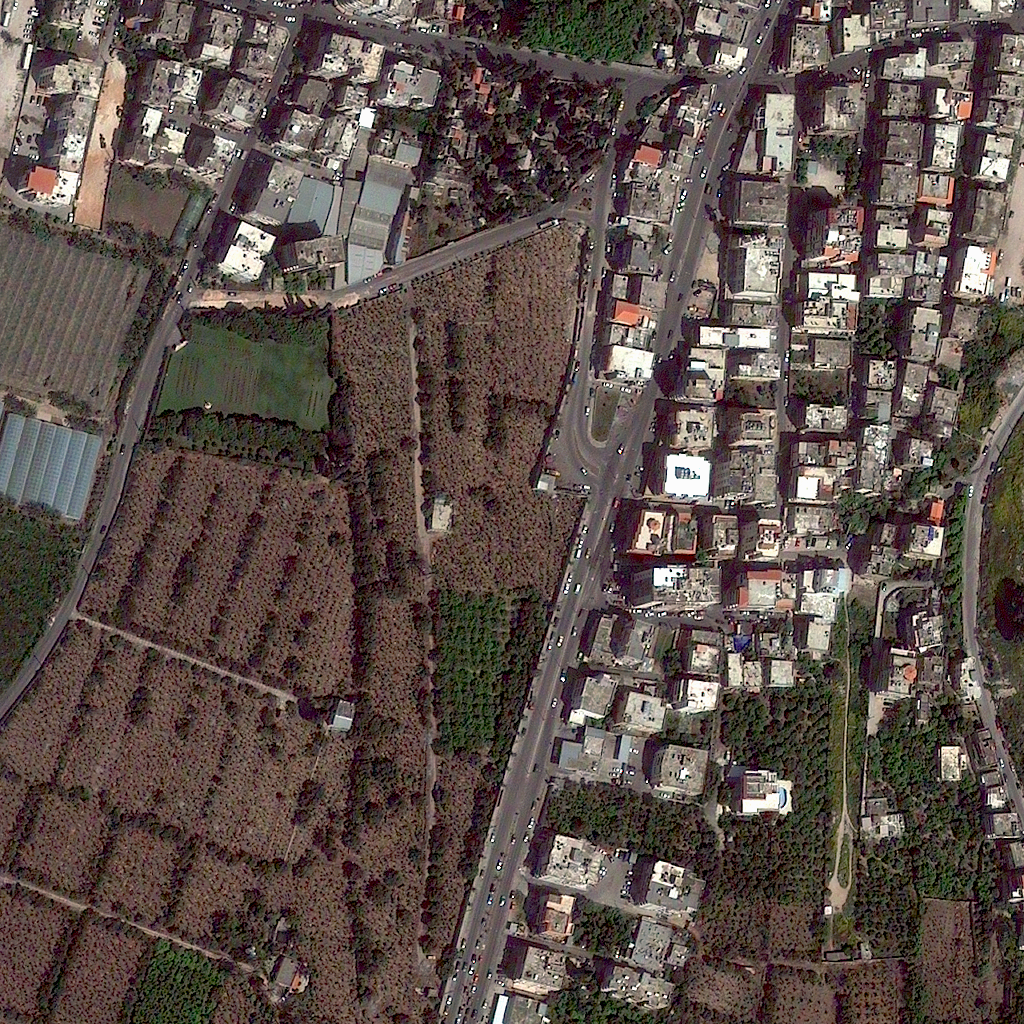}}
\subfloat[Saida Instances Mask]{\includegraphics[width = 1.5in]{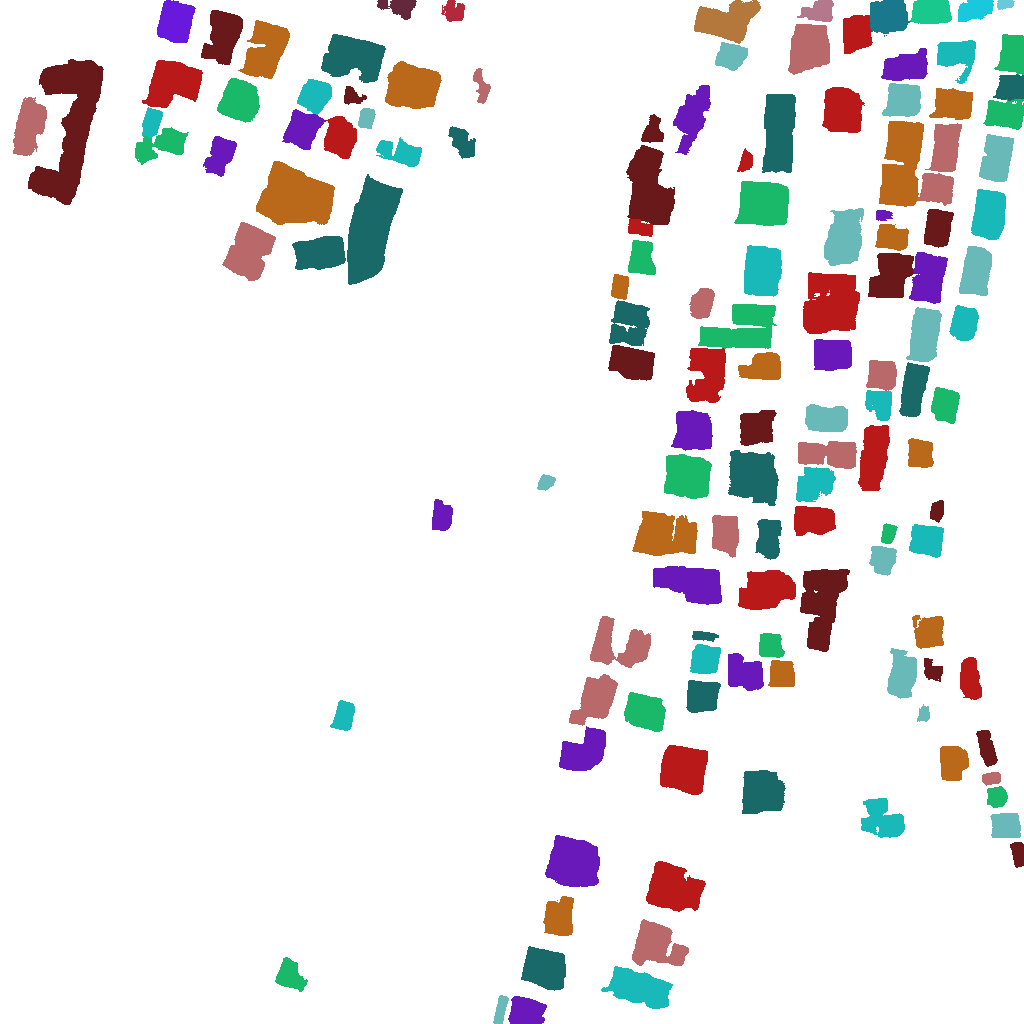}}\\
\subfloat[Tripoli Sample]{\includegraphics[width = 1.5in]{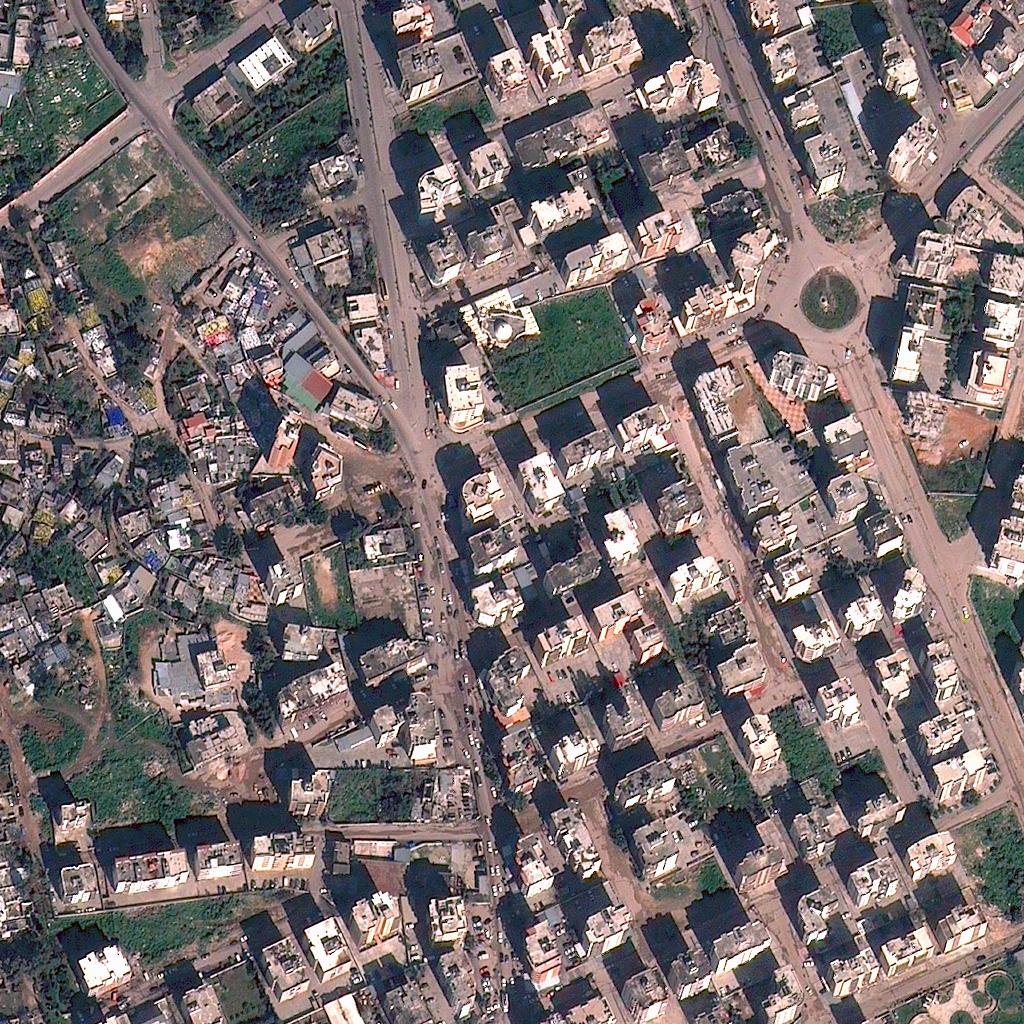}}
\subfloat[Tripoli Instances Mask]{\includegraphics[width = 1.5in]{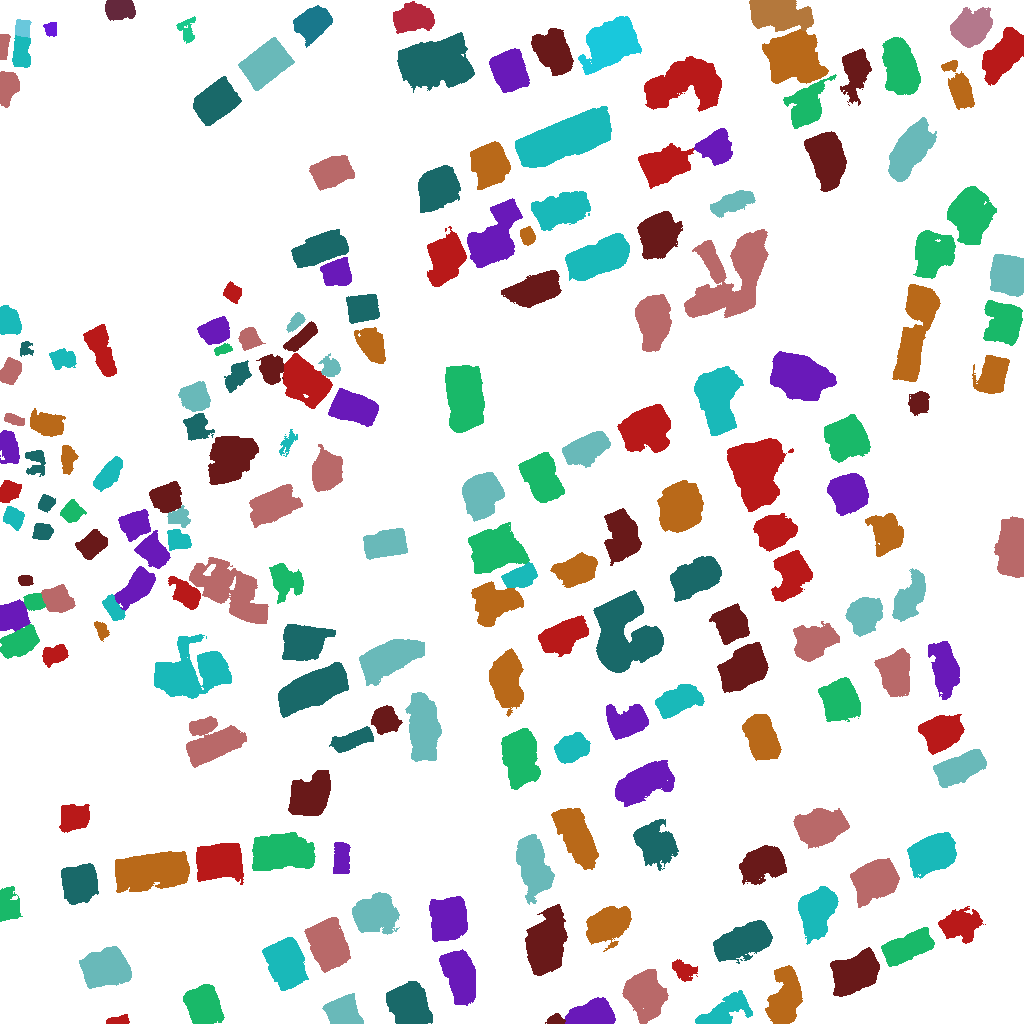}} \\
\subfloat[Byblos Sample]{\includegraphics[width = 1.5in]{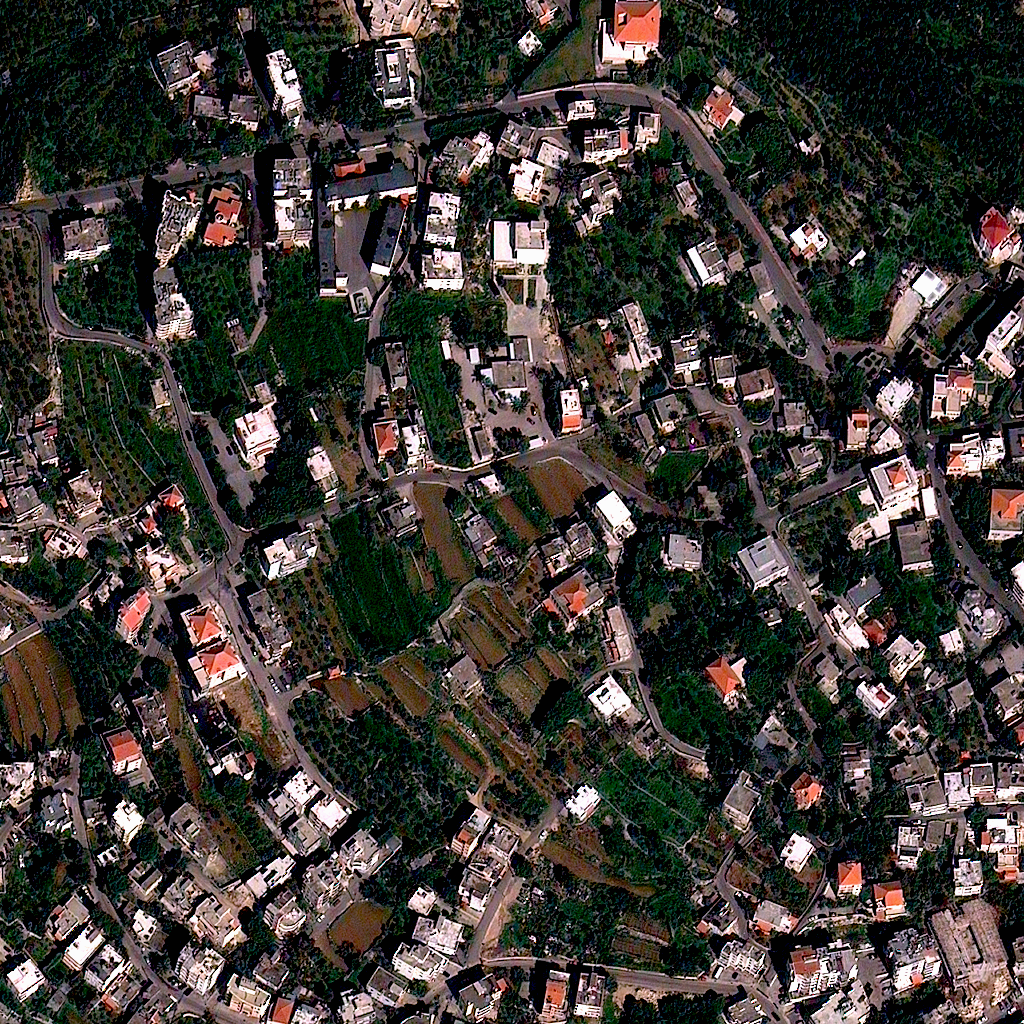}}
\subfloat[Byblos Instances Mask]{\includegraphics[width = 1.5in]{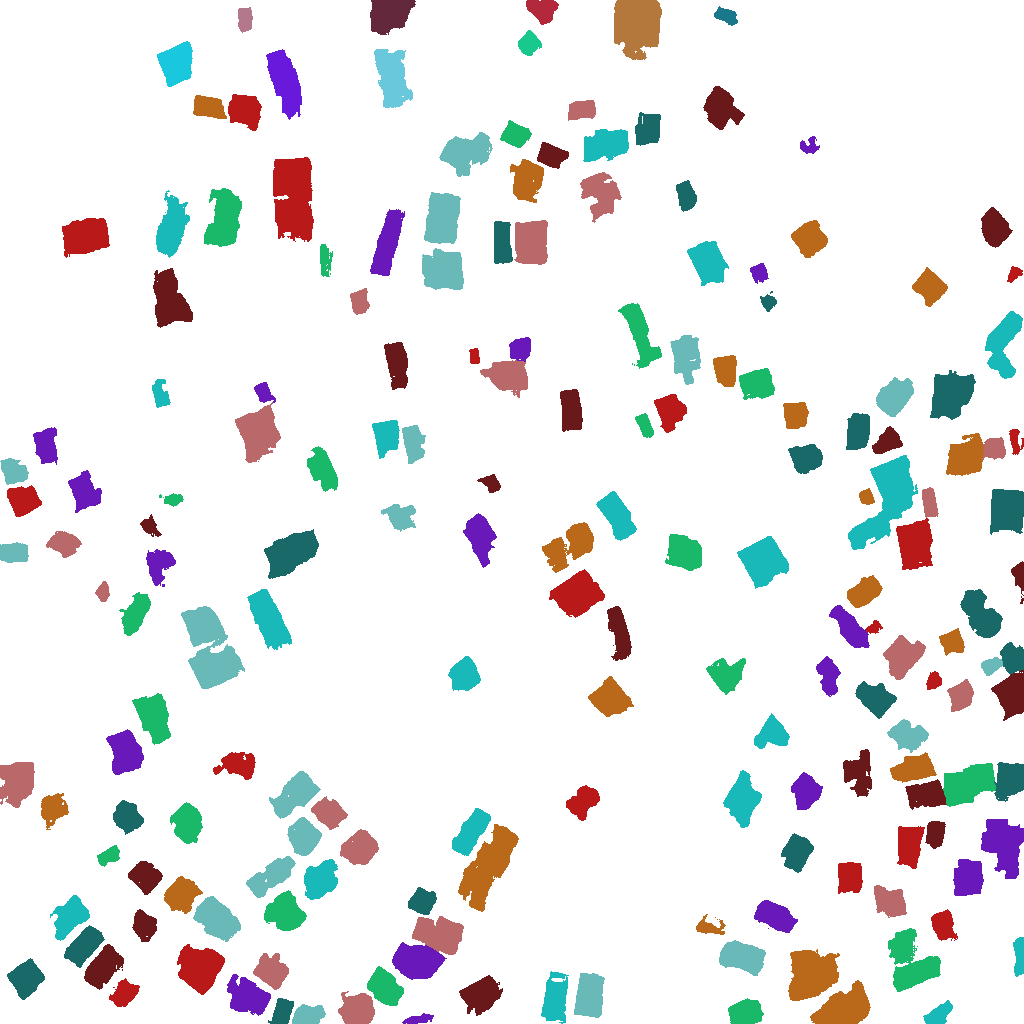}} \\
\end{center} 
\caption{Sample images and their corresponding instance segmentation mask using multi-class inference. The buildings are represented in different colors to emphasize the fact that each instance has its own identifier.}
\label{inf4} 
\end{figure}

To further test the proposed pipeline, we sampled various tiles from different regions of Lebanon, including Beirut, Saida, Byblos and Tripoli, for which we ran the proposed pipeline and calculated the F-score results as shown in Table \ref{tb4}. In Figure \ref{inf4}, we qualitatively show the instance segmentation results for those different regions. Each building is presented in a different color to emphasize the fact that we are conducting instance and not semantic segmentation.

The results in Table \ref{tb4} show that the muti-class model approach provides up to 14\% increase in F-score. This gain is achieved by the better ability of the multi-class model to separate very close buildings. The Nadir angle for very dense areas with tall buildings, such as Beirut, is a major factor that affects the performance of the model, which explains the drop in the F-score in this region. In future work, we plan to tackle this issue and provide practical workarounds. For regions with proper urban planning, such as Saida, our model provides an outstanding F-score of 81.5\%.

\begin{table}[h]
\begin{center}
\begin{tabular}{c|c|c} \toprule 
     {Region} & \shortstack{single-class \\ F-score(\%)} & \shortstack{multi-class \\ F-score(\%)}\\ \midrule
     {Saida} & {67.40} & {81.56}\\
     {Tripoli} & {56.15} & {70.20}\\
     {Byblos} & {56.68} & {67.03}\\
     {Beirut} & {49.07} & {55.10}\\\bottomrule
\end{tabular}
\end{center}
\caption{A comparison between single-class and multi-class F-score results on images from different regions.}
\label{tb4}
\end{table}

Furthermore, to further investigate the performance per image for the single- vs multi-class methods, we draw the violin plots in Figure \ref{violin} to compare the probability density for the True Positive (TP), False Positive (FP), and False Negative (FN) predictions per image over the test dataset. The dashed line represents the median of each distribution and the dotted lines represent the Inter-Quartile Ranges (IQR). The IQR represents the `Middle 50\%' distribution of the data, so it provides a sense of the concentration of values. Figure \ref{violin} shows that the median and IQR boundaries of the True Positive are larger for the multi-class model. Moreover, the median and IQR boundaries of False Positive and False Negative plots are smaller for multi-class model. The data distribution shown in the violin plot emphasizes the gains of the proposed model at test time.

\section{Conclusion}
\label{conclusion}
In this paper, we introduced multi-class segmentation model to extract building footprints from satellite imagery at a national level. We also presented a demo case study for Lebanon. The proposed pipeline has impact and use cases on several technologies, such as the national solar potential map, the planning of utilities at the local government level, urban agglomeration, modeling air pollution, modeling solid waste, and many others.


 
 \section{Acknowledgment}
This project was partially funded through the nVidia GPU Grant. The authors thank Egn. Hanaa Makki for preparing the map illustration shown in Figure 1.

\bibliographystyle{IEEEbib}
\bibliography{references}

\end{document}